\documentclass[sigconf]{acmart}
\usepackage{todonotes}
\usepackage{mdframed}
\usepackage{algorithmicx}
\usepackage{natbib}
\usepackage{multirow}
\usepackage[normalem]{ulem}
\usepackage[noend]{algpseudocode}
\usepackage{flushend}
\usepackage{graphicx}
\usepackage{multicol}
\usepackage{subcaption}

\setcopyright{none}

\begin{document}

\settopmatter{printacmref=false} 
\renewcommand\footnotetextcopyrightpermission[1]{} 
\pagestyle{plain} 

\title{Augmented Convolutional LSTMs for Generation of High-Resolution Climate Change Projections}

\author{Nidhin Harilal}
\affiliation{%
  \institution{IIT Gandhinagar}}
\email{nidhin.harilal@iitgn.ac.in}

\author{Udit Bhatia}
\affiliation{
  \institution{IIT Gandhinagar}}
\email{bhatia.u@iitgn.ac.in}

\author{Mayank Singh}
\affiliation{
  \institution{IIT Gandhinagar}}
\email{singh.mayank@iitgn.ac.in}

\renewcommand{\shortauthors}{Harilal, et al.}

\begin{abstract}
Projection of changes in extreme indices of climate variables such as temperature and precipitation are critical to assess the potential impacts of climate change on human-made and natural systems, including critical infrastructures and ecosystems. While impact assessment and adaptation planning rely on high-resolution projections (typically in the order of a few kilometers), state-of-the-art Earth System Models (ESMs) are available at spatial resolutions of few hundreds of kilometers. Current solutions to obtain high-resolution projections of ESMs include downscaling approaches that consider the information at a coarse-scale to make predictions at local scales. Complex and non-linear interdependence among local climate variables (e.g., temperature and precipitation) and large-scale predictors (e.g., pressure fields) motivate the use of neural network-based super-resolution architectures. In this work, we present auxiliary variables informed spatio-temporal neural architecture for statistical downscaling. The current study performs daily downscaling of precipitation variable from an ESM output at 1.15 degrees (~115 km) to 0.25 degrees (25 km) over the world's most climatically diversified country, India. We showcase significant improvement gain against three popular state-of-the-art baselines with a better ability to predict extreme events.  To facilitate reproducible research, we make available all the codes, processed datasets, and trained models in the public domain.
\end{abstract}




\maketitle
\section{Introduction}

The last few decades have witnessed record-breaking climate and weather-related extremes across the globe~\cite{prein2017future,zhang2013attributing}. We are witnessing urgent calls for substantial new investments in climate modeling to increase the precision, accuracy, reliability, and resolution at which information is made available to the stakeholders~\cite{dessai2009we, schmidt2010real}. To model earth's past, present, and future climate, Earth System Models are used for simulate the interactions of atmosphere, land, ocean, ice and biosphere to estimate the state of local and regional climate under various conditions. 
Earth system models (ESMs) integrate the interactions of atmosphere, ocean, land, ice, and biosphere to estimate the state of regional and global climate under a wide variety of conditions \cite{arora2013carbon}. Despite the increasing complexity and inclusion of more sophisticated schemes to account for interactions across various constituting components, ESMs suffer from three broad sources of uncertainties: (i) knowledge gaps in understanding of coupled natural-human systems which could results in different Representative Concentration Pathways (RCPs), (ii) lack of understanding of physics of climate system which is encapsulated through parametric differences in Multiple Model Ensembles (MMEs), and (iii) intrinsic variability which is captured through multiple initial condition ensembles \cite{kumar2018intercomparison}. To characterize these uncertainties, ESMs are often executed with multiple initial conditions, multiple ensemble mode with different RCPs making impact relevant insight discovery from ESM  outputs a ``big data'' challenge \cite{faghmous2014big}.  Owing to ESMs high-computation requirements, their projections are limited to a very coarse resolution  (typically  100--150 km), leading to complications in assessing the regional impacts \cite{wilby2010robust,dessai2009climate}.  While there has been an emphasis on multiple model ensembles for assessing climate change impacts on precipitation and temperature, researchers believe that ignoring the contribution of internal variability could result in underestimation of statistics of extremes, which in turn can lead to maladaptation~\cite{bhatia2019precipitation}. However, the development of design relevant intensity-duration and frequency curves require climate projections at high resolution. Hence, generating high-resolution data products that have high credibility is pivotal to inform adaptation strategies in the backdrop of climate change.  

\subsection{The Downscaling Approaches}
Climate scientists use downscaling techniques to generate climate projections at higher spatial resolution. These techniques are broadly classified into two categories, namely dynamical and statisti`cal downscaling.  Dynamical downscaling embeds the sub-grid processes within boundary conditions of coarse resolution ESM grids. Examples of sub-grid processes include cloud physics, radiation, soil characteristics, and hydrological processes.  While dynamical downscaling is useful for simulating extreme precipitation events and localized phenomena such as convective precipitations, the generated outputs are highly sensitive to the boundary conditions~\cite{xue2014review}. 
High computational requirements and limited spatial scalability limit the usage in downscaling multiple ensembles, and multiple initial conditions runs of ESMs.
On the other hand, statistical downscaling (hereafter \textit{`SD'}) attempts to learn the statistical relationship between ESM output (at coarse resolution) with high-resolution observations such as observed or remotely sensed precipitation. With an underlying assumption of space and time stationarity, the above relationship helps in generating higher resolution outputs of coarse resolution ESM projections. 

\subsection{Climatically Diversified Indian Sub-continent}

Indian subcontinent comprises a wide range of weather conditions and six climate subtypes ranging from tundra and glaciers in the north to deserts in the west, and humid tropical regions in the southwest.  
Indian subcontinent comprises of wide range and highly disparate micro-climate systems, making it one of the most climatically diverse countries on the globe \cite{brenkert2005modeling}. Despite the stark heterogeneity, these geographical and geological features work in tandem to drive Indian Summer Monsoon Rainfall (or ISMR) \cite{kulkarni2009spatial, clift2008asian}.  India receives 80\% of its annual rainfall during the southwest summer monsoon season. While monsoon systems are reliable at annual scales as a consequence of seasonal heating of the land, even small percentage variations \cite{mishra2012prominent} can have dramatic impacts on food and economic security of the nation \cite{gadgil2006indian}. In addition to regional and local correlations in space and time, ISMR is influenced by the coupled ocean and atmospheric phenomena over the Indian Ocean, Pacific decadal oscillations, and Atlantic Ocean sea surface temperature making it a complicated system to predict. 
Given the utmost importance of ISMR in the Indian Climatic system, we have trained the proposed model separately for monsoon and non-monsoon seasons in the context of SD.

\subsection{Spatio-temporal Teleconnections}
In atmospheric sciences, climate variables and anomalies often exhibit long-term dependence among spatially non-contiguous regions, referred to as \textit{``teleconnections''}~\cite{adarsh2019links}. Analogously, it is widely accepted that climate variability is scale-invariant and exhibit long-term memory in time~\cite{yuan2013long}. Here, long-term memory implies that the present state of the system influences the future states.   Recent advancements in deep learning literature result in several SD architectures such as Convolution Neural Networks (CNNs)~\cite{vandal2017deepsd}, Residual Dense Block (RDB) \cite{zhang2018residual} and Long Short Term Memory (LSTMs)~\cite{misra2018statistical} capturing spatial and temporal dependencies, respectively.

However, to the best of our knowledge, we do not find any SD technique that captures both spatio-temporal dependencies simultaneously. Moreover, we showcase that both CNN-based approaches~\cite{vandal2017deepsd} and RDB based approaches that generates high-resolution projections of ESM using a single image does not perform well when applied to real ESM data. 

\subsection{Augmented Convolutional LSTMs}
In this paper, we address the major limitation of SD techniques in capturing spatio-temporal dependencies simultaneously.  We discuss a framework to downscale multiple initial condition ensembles of Community Earth System Model Large Ensemble (CESM-LENS) project~\cite{kay2015community} using  Recurrent Convolutional LSTM. Besides, we propose a research direction to augment multiple state variables in addition to station elevation data by including seven physics guided auxiliary variables.

\subsection{Key Contributions}
The key contributions are as follows:
\begin{itemize}
    \item We present a Recurrent Convolutional LSTM based Super-resolution approach towards statistical downscaling of climatic data from coarse-resolution ESM to fine-resolution Observation data, which accounts for spatial and temporal dependence in space and time between target variable (precipitation in the present case) and auxiliary variables. We propose a  way to include multiple state variables in addition to station elevation data by including seven physics guided auxiliary variables.
    \item We address the major limitation of state-of-the-art deep learning-based super-resolution architectures that use a coarse resolution version of the same image to generate finer resolution outputs. Moreover, in reality, ESMs are expected to capture statistics of atmospheric variables, especially at decadal and interdecadal scales. Hence, day-to-day mapping, as proposed by earlier models, is of limited to no use to multiple downscale models and multiple ensembles of ESMs.    
\end{itemize}


\subsection{Organization of the Paper:}
We organize the rest of the paper as follows. Section~\ref{sec:dataset} outlines data used in present research along with the rationale of choice of covariates. Section~\ref{sec:limit} discusses the related work in the area of SD along with an overview of state-of-the-art convolution neural network and LSTM based approaches for SD and their key limitations. Section~\ref{sec:method} presents ConvLSTM for SD. Section~\ref{sec:exp} describes the experimental settings. In Section~\ref{sec:results}, we compare our results with the current state-of-the-art ResLap \cite{reslap}, DeepSD~\cite{vandal2017deepsd} and widely used quantile mapping technique~\cite{cannon2011quantile}. In Section~\ref{sec:con}, we briefly discuss results, limitations, and scalability potential of our work to generate high-resolution outputs that can be leveraged by stakeholders and policymakers for design and adaptation planning. 

\section{Datasets}
\label{sec:dataset}

Climate modeling relies on the numerical solution of the fundamental equations for atmospheric motion, i.e., conservation of mass, energy, and momentum, and equation of state, which are expressed as simultaneous non-linear dynamical equations~\cite{hansen1983efficient}. The solution of these equations typically yields state variables including temperature, humidity, atmospheric pressure, wind velocities in three directions (zonal, meridional, and vertical velocities), and precipitation. However, running the climate models or ESMs at higher temporal resolutions results in higher computation cost since the relationship between spatial resolution and model execution time is non-linear; hence these models run at the resolution of a 100--150 km resulting in too coarse resolution for local and regional adaptation. In this paper, we use coarse resolution precipitation outputs from the NCAR Community Earth System Model (NCAR CESM1 CAM5) obtained from the archives of the Climate Modeling Intercomparison Project 5~\cite{lauritzen2018ncar,taylor2012overview}. To establish the relationship between coarse resolution model output and observations at high resolution, we use the high-resolution spatial resolution ($0.25^o\times 0.25^o)$ available for 110 years over the Indian mainland~\cite{pai2014development}. As precipitation is also dependent upon temperature, humidity, atmospheric pressure, and wind velocities, we use these variables as potential covariates in addition to coarse resolution precipitation. Since observations for these covariates is not available at the gauge level, we use the outputs from the national Centers for Environmental Prediction-National Center for Atmospheric Research (NCEP-NCAR) global reanalysis project~\cite{kalnay1996ncep}.  Furthermore, the correlation between precipitation patterns and topography suggests that type (rain or snow), intensity, and duration of precipitation is significantly controlled by elevation characteristics. Hence, we use the topographical information from NASA's Shuttle Radar Topography Mission (SRTM), which is available at 90 meters. As outlined in~\cite{vandal2017deepsd}, we use the image construct to combine coarse resolution precipitation with covariates and elevation data (jointly referred to as \textit{`auxiliary variables'}), resulting in a 7-channel input image. One advantage of retaining image-based construct over other machine learning-based approaches of SD is that it preserves the spatial dependence structure. Section~\ref{sec:dataset_prep} presents detailed description of datasets with experimental settings. 

\begin{figure}[!tbh]
    \centering
    \includegraphics[width=0.5\textwidth]{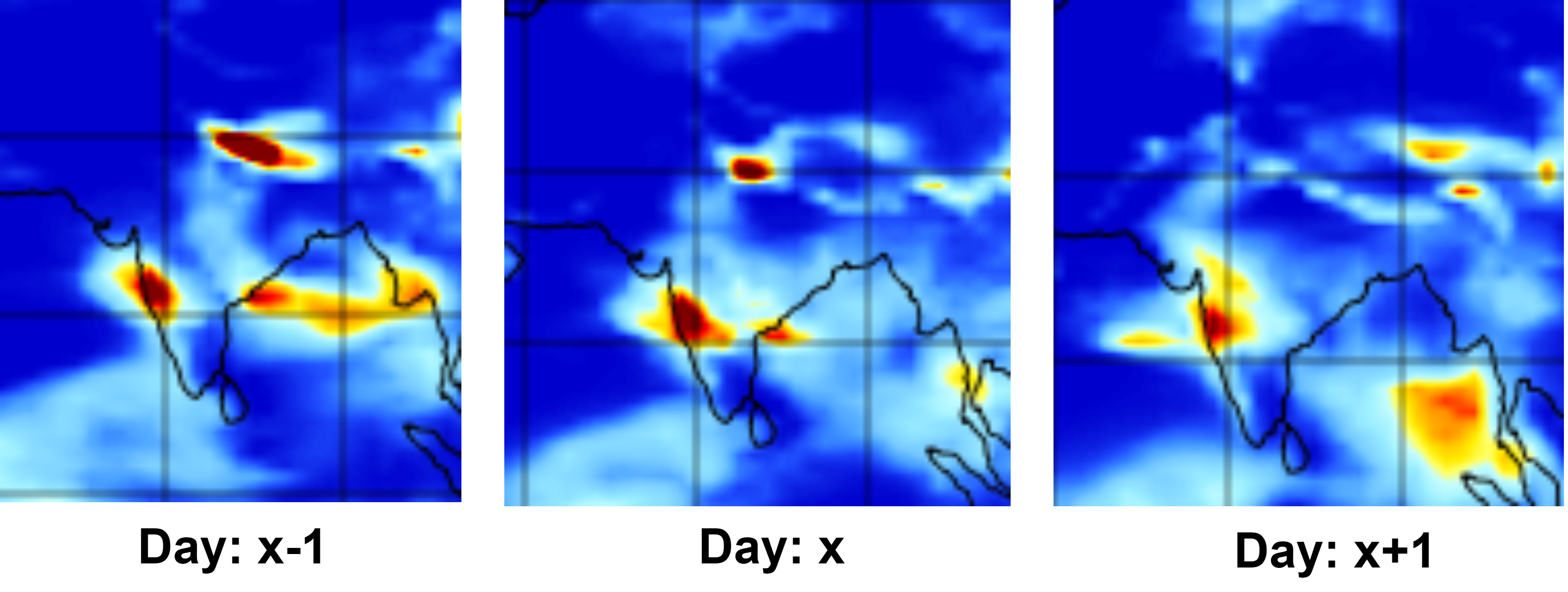}
    \caption{\textbf{(Best viewed in color) An example of temporal dependence of precipitation over a span of three days. Red color represents regions of high precipitation, whereas Blue color represents regions with low precipitation. 
    } }
    \label{fig:temp}
\end{figure}

\section{Related Work}
\label{sec:limit}

As discussed before, SD is a technique for obtaining high-resolution climate and/or climate change information at adaptation relevant scales. SD has been applied to coarse resolution ESM output for years with a rich body of work existing in this field.  SD is typically viewed as a problem of regression with the objective of identifying the optimal transfer function between ESM inputs and observations. Both linear and non-linear approaches are used to establish the relationship between auxiliary variables and the target variable of interest. Popular approaches among others include linear regression methods~\cite{wilby1997downscaling}, Automated Statistical Downscaling (ASD)~\cite{guo2012prediction}, relevance vector machines~\cite{ghosh2008statistical}, zero-shot super-resolution architecture \cite{shocher2018zero}, quantile regression neural network~\cite{cannon2011quantile}, and multivariate linear regression approach~\cite{chen2010statistical}. Another widely used approach is quantile mapping methods~\cite{cannon2015bias}, which correct systematic biases in the distribution of climate variables. While these approaches are sufficient to capture mean statistics of climate variables, it has been found that these approaches fail in capturing extremes. Moreover, while quantile mapping methods have gained wider popularity, these tend to corrupt the trends projected by models artificially~\cite{cannon2015bias}. 

Given the spatiotemporal nature and underlying non-linear behavior of the climate system, there is growing interest in the adaptation of deep learning techniques based super-resolution architectures to statistical downscaling.
\textcolor{black}{Vandal \emph{et al.}~\cite{deepsd} proposed DeepSD which considered the complex precipitation data as a single image. DeepSD~\cite{deepsd} consists of convolution neural networks based super-resolution architecture SRCNN~\cite{srcnn} to capture the spatial dependencies. Inspired by this, other researchers have proposed ResLap~\cite{reslap} which uses laplacian pyramid based super-resolution network~\cite{LapSRN} for further improvement in the quality of the generated climate change projections. ResLap~\cite{reslap} extracts hierarchical features from the original precipitation projection and used a series of transposed convolution networks to upscale the low-resolution projection to the target scale.}
On other hand, given the long short term memory processes that are inherent to the system, researchers have also used recurrent and vanilla Long Short Term Memory (LSTM)~\cite{sudeshna} for climate downscaling. Moreover, in the context of SD using super-resolution construct, authors have envisioned the problem of SD as that of a single image Super-resolution. However, this approach has two major limitations. First, ESMs are designed to capture the statistics of climate variables, and there is no concept of one-to-one mapping with observations. Hence, single image Super-resolution can yield superfluous results when applied to real ESMs. Secondly, while both DeepSD\cite{deepsd} and ResLap\cite{reslap} accounts for dependencies in space, temporal dependencies are altogether ignored, which are ubiquitous in climate and weather patterns, as shown in \textbf{Fig~\ref{fig:temp}}. 
To capture temporal dependence,~\cite{sudeshna} used Long-Short Term Memory (LSTM) recurrent neural network. LSTMS are commonly used for processing sequential inputs and are preferred over vanilla RNNs as they solve the problem of vanishing gradients to some extent. However, one of the essential requirements for LSTMs is that it requires a one-dimensional data stream. So, in order to fit this with the problem, authors flattened the image to form a single-dimensional array and passed it to LSTMs, resulting in loss of spatial structure. To our knowledge, little work has been attempted to explicitly capture dependency in space and time in the context of SD.

\section{Methodology}\label{Methodology}
\label{sec:method}
In this section, we present a detailed discussion of our proposed SD methodology. The proposed deep architecture addresses the problems described in the previous section. 

\subsection{Convolutional LSTM Networks for SD}

We envision the problem of SD as learning the transfer function between high-resolution observations and coarse resolution ESM outputs using a spatio-temporal sequence of state variables as input. By combining fully connected LSTMs with convolutions, we build an end-to-end trainable convolutional LSTM SR (ConvLSTM-SR) model for SD problem that combines convolutional LSTMs~\cite{xingjian2015convolutional} with super-resolution block (discussed next). 

\subsection{Model Description}
Consider $X_t$ representing low resolution spatial data at $t^{th}$ day consisting of seven channels, each representing one climate variable. Each $X_t$ represents an ``image'' of size 7$\times$N$\times$M, where last two dimensions representing spatial dimensions. Consider $\mathcal{X}_t$ representing previous $T$ low resolution spatial data points available on $t^{th}$ day: 
\begin{equation}
    \mathcal{X}_t = [X_{t-T}, X_{t-(T-1)}, X_{t- (T-2)}, \ldots , X_{t-1}, X_{t}] 
\end{equation}

We define convolutional LSTM, $ConvLSTM(\eta,k)$, where $\eta$ and $k$ represent number of filters and kernel dimensions, respectively. Convolutional LSTM handles spatio-temporal data by considering cell outputs $C_1,\ldots,C_t$, hidden states $H_1,\ldots,H_t$, and gates, $i_t$ (input), $f_t$(forget) and $o_t$ (output) as 3D tensors. \textcolor{black}{The equation below~\cite{convlstm}} illustrates operations at different gates with $\ast$ and $\circ$ representing convolution and element-wise matrix multiplication operations, respectively.
\vskip -0.3cm

\begin{equation}
  i_t = Sigmoid(W_{i}\ast \mathcal{X}_t + R_{i} \ast h_{t-1} + \rho_{i} \circ C_{t-1}+b_i)
\end{equation}
\vskip -0.1cm
\begin{equation}
  f_t = Sigmoid(W_{f}\ast \mathcal{X}_t + R_{f} \ast h_{t-1} + \rho_{f} \circ C_{t-1}+b_f)
\end{equation}
\vskip -0.1cm
\begin{equation}
  o_t = Sigmoid(W_{o}\ast \mathcal{X}_t + R_{o} \ast h_{t-1} + \rho_{o} \circ C_{t} +b_o)
\end{equation}
\vskip -0.1cm
\begin{equation}
  C_t = f_t \circ c_{t-1} + i_t \circ Tanh(W_{c}\mathcal{X}_t + R_{c}h_{t-1}+b_c)
\end{equation}
\vskip -0.1cm
\begin{equation}
  h_t = o_t \circ Tanh(c_t)
\end{equation}

\begin{figure}
    \centering
    \includegraphics[width=0.5\textwidth]{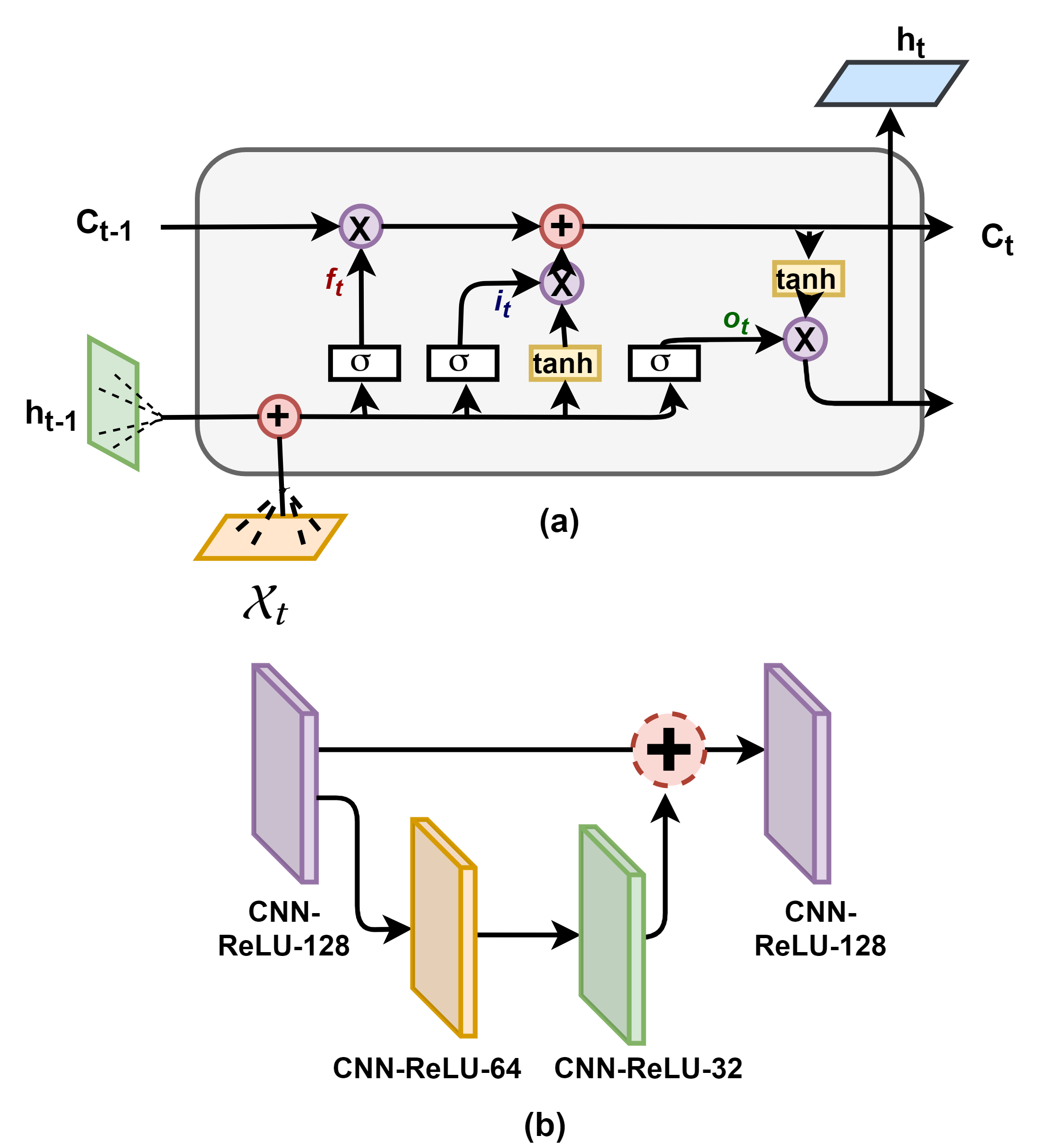}
    \caption{\textbf{(a) A detailed Representation of ConvLSTM block. (b) Representation of Super-resolution (SR) block.\textcolor{black}{ The labels of each block follows the convention of name of the layer followed by its activation and an integer represents the number of kernels used in that particular block.}}}
    \label{fig:convLSTM}
    \label{fig:SR-block-architecture}
\end{figure}

In the above equations, $W_i$, $W_f$, $W_o$ and $W_c$ $\in$ $\mathbb{R}^{\eta,k,k}$ represent input weights, $R_i$, $R_f$, $R_o$ and $R_c$ $\in$ $\mathbb{R}^{\eta,k,k}$ represent recurrent weights, $\rho_{i}$,  $\rho_{f}$, and $\rho_{o}$ $\in$ $\mathbb{R}^{k}$ represent peephole weights and $b_i$, $b_f$, $b_o$ and $b_c$ $\in$ $\mathbb{R}^{k}$ represent bias weights.   \textbf{Fig~\ref{fig:convLSTM}} shows the detailed description of ConvLSTM block. 

We propose a stacked ConvLSTM architecture, comprising three stacked ConvLSTM blocks and a super-resolution (SR) block. The first ConvLSTM block attempts to encode the temporal information present in the sequence while maintaining the spatial dependencies between them with total $\eta_1$ filters of $k_1 \times k_1$ dimensions.
\begin{equation}
    F_1(\mathcal{X}) = ConvLSTM (\eta_1,k_1)(\mathcal{X})
\end{equation}

Similarly, second ConvLSTM block comprises $\eta_2$ filters each with $k_2 \times k_2$ dimensions.

\begin{equation}
    F_2(\mathcal{X}) = ConvLSTM (\eta_2,k_2)(F_1(\mathcal{X}))
\end{equation}

The last ConvLSTM block comprises $\eta_3$ filters each with $k_3 \times k_3$ dimensions.

\begin{equation}
    F_3(\mathcal{X}) = ConvLSTM (\eta_3,k_3)(F_2(\mathcal{X}))
\end{equation}

In our experiments, we choose $N$ and $M$ as 129 and 135. We choose $\eta_1$, $\eta_2$, $\eta_3$ as 32, 16, and 16, respectively. Similarly, we choose $k_1$, $k_2$, and $k_3$  as 9, 5, 3, respectively. The size of the resultant output tensor after three stacked ConvLSTM operations is 16$\times$129$\times$135. The recurrent nature associated with the ConvLSTM helps in utilizing the temporal dependencies associated with data. On the other hand, the convolutional operations inside the cell help in retaining the spatial correlation. Therefore, The output tensor encodes temporal as well as the spatial dependencies associated with the data.


The super-resolution (SR) block (described in \textbf{Fig~\ref{fig:SR-block-architecture}}) increases the resolution of high dimensional feature data obtained from the preceding ConvLSTM layers. In our experiments, SR block consist of a six stacked deep Convolutional layers with skip connections in between them. \textcolor{black}{Each convolution layer has Relu activation~\cite{relu}.} Below shows operation sequence inside the Super-resolution (SR) block. $W^\prime_{1}$, $W^\prime_{2}$, ,$W^\prime_{3}$ and $W^\prime_{4}$ represent filters of size 16$\times$9$\times$9, 128$\times$5$\times$5 and 64$\times$3$\times$3, 32$\times$3$\times$3 respectively. $B^\prime_1$, $B^\prime_2$, and $B^\prime_3$ represent bias weights.


\begin{equation}
F^\prime_1(X) = W^\prime_{1} \ast X + B^\prime_1\\
\end{equation}

\begin{equation}
F^\prime_2(X) = ReLU(W^\prime_{2} \ast F^\prime_1(X) + B^\prime_2)\\
\end{equation}

\begin{equation}
F^\prime_3(X) = ReLU(W^\prime_{3} \ast F^\prime_2(X) + B^\prime_3)\\
\end{equation}

\begin{equation}
F^\prime_4(X) = CONCAT(F^\prime_1(X), F^\prime_3(X))\\
\end{equation}

\begin{equation}
F^\prime_5(X) = W^\prime_{4} \ast X + B^\prime_4
\end{equation}

As described above, the high-resolution feature data processed produced from the last ConvLSTM layer is passed to the first SR block. 
\begin{equation}
    F_4(\mathcal{X}) = SRBlock(F_3(\mathcal{X}))
\end{equation}

The output from the first SR block is further passed through another SR block for doubling the current resolution. 

\begin{equation}
    F_5(\mathcal{X}) = SRBlock(F_4(\mathcal{X}))
\end{equation}

\begin{equation}
    F_6(\mathcal{X}) = ReLU(W^{\prime\prime}_1 \ast F_5(\mathcal{X}) + B^{\prime\prime}_1)
    \label{eg:f6}
\end{equation}

\begin{equation}
    F_7(\mathcal{X}) = W^{\prime\prime}_2 \ast F_6(\mathcal{X}) + B^{\prime\prime}_2
    \label{eg:f7}
\end{equation}

\begin{figure*}[!tbh]
        \includegraphics[width=2\columnwidth]{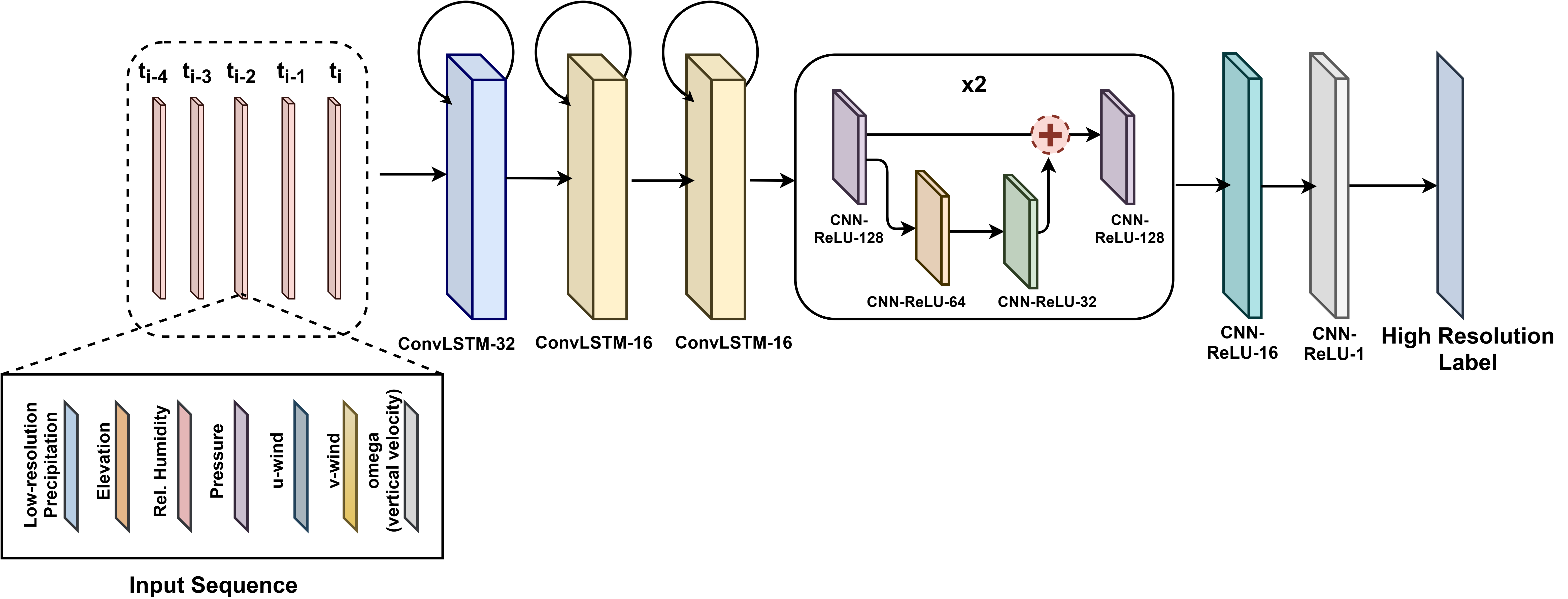} 
        \caption[a floating figure]{\textbf{Proposed Convolutional LSTM based model augmented with Super-resolution block. On left, the representation of all the 7 climatic variables (Precipitation, Elevation, Relative-Humidity, Pressure, 3 Wind components) as channels is shown. \textcolor{black}{The numbers mentioned in the labels of each block followed by its name and activation represents the number of kernels used in that particular block.} }}
        \label{fig:architecture}
\end{figure*}

In our experiments ~\eqref{eg:f6} and ~\eqref{eg:f7}, $W^{\prime\prime}_1$ and $W^{\prime\prime}_2$ represent filters of size 128$\times$1$\times$1 and 16$\times$3$\times$3, respectively. $B^{\prime\prime}_1$ and $B^{\prime\prime}_2$ represent bias weights. The overall architecture is shown in \textbf{Fig~\ref{fig:architecture}}.


\section{Experiments}
\label{sec:exp}
This section describes the dataset and the experimental settings.

\subsection{Dataset}
\label{sec:dataset_prep}


Owing to the diversity in the Indian precipitation profile, we categorize months into two periods: (i) Non-monsoon (October -- April) and (ii) Monsoon months (May -- September). \textcolor{black}{Since the spatial resolution of the ESM(1.15°), the reanalysis state variables(2°) and observations(0.25°) are different, therefore, we interpolate both the ESM and the reanalysis state variables to match the resolution of the observation.}
We train our proposed model for these periods separately. In addition, each variable is normalized for better representation. Except for precipitation, all climate variables are normalized between 0 and 1. Precipitation is normalized between 0 and 50.  This differential normalization scheme explicitly focuses the model on the precipitation variable compared to other auxiliary variables. 
\begin{table}[t]
    \centering
    \caption{\textbf{Dataset statistics}}
    \begin{tabular}{c|c|c}
    \toprule
    \multirow{5}{*}{\rotatebox[origin=c]{90}{\centering\textbf{Overall}}}& Year-range &  1948--2005 \\
    &Climate variables & 7\\
    &Total instances & 21,170\\
    &Input shape (interpolated) & $7\times 129 \times 135$\\
    &Output Shape &$1\times 129 \times 135$\\
    \hline
     \multirow{2}{*}{\rotatebox[origin=c]{90}{\centering\textbf{Test}}}&Year-range&2000--2005\\
     &Instances&2,190\\
     \hline
    \multirow{2}{*}{\rotatebox[origin=c]{90}{\centering\textbf{Train\:}}}&
    Year-range&1948--1999\\
     &Instances&20,075\\
    \bottomrule   
    \end{tabular}

    \label{tab:dataset}
\end{table}

We, further, split data into train and test classes. The training data comprises years between 1948--1999, whereas test data comprises years between 2000--2005. 

\subsection{Experimental settings}

\subsubsection{Parameter Training}
The proposed model generates fine-grained precipitation projections leveraging previous five days coarse-grained climatic projections. 
We incorporate recurrent dropouts of $0.2$ to the stacked ConvLSTM layers along with a dropout of $0.1$ between ConvLSTM layers. Additionally, we keep regularization with weight decay of value $0.02$ in the ConvLSTM layers. The RMSE loss is optimized using Adam optimizer~\cite{kingma2014adam}. Besides,  we keep an adaptive learning rate ($lr$) with an initial value of $0.0003$. The subsequent rates follow the update equation, $lr = lr*\alpha$, where $\alpha=0.2$. The models for both monsoon and non-monsoon periods have an identical set of parameters. Both the models are trained for 1500 epochs with the mini-batch size as 15.
Models were built and trained on Tensorflow \cite{abadi2016tensorflow}. For training, two NVIDIA RTX 2080-Ti GPUs were harnessed by independently training the two models (monsoon and non-monsoon) on each of them. 

%

\subsection{Baselines}
We compare our proposed model with two state-of-the-art baselines, ResLap~\cite{zhang2018residual, reslap},  DeepSD~\cite{vandal2017deepsd} and Quantile Mapping Approach~\cite{cannon2015bias}.
\subsubsection{ResLap\cite{reslap}}
\textcolor{black}{It uses a Residual Dense Block (RDB) on top of Laplacian pyramid based super-resolution network~\cite{LapSRN} for generating high resolution climate change projections. ResLap~\cite{reslap} extracts hierarchical features at different levels from the original precipitation projection. It then uses a series of transposed convolution networks to upscale the low-resolution projection to the target scale.}
\subsubsection{DeepSD~\cite{vandal2017deepsd}} \textcolor{black}{It uses a series of stacked convolutional neural network-based super-resolution architecture SRCNN~\cite{srcnn}. Each of the stacked SRCNN~\cite{srcnn} module in DeepSD~\cite{deepsd} is trained independently to generate climate change projection at different resolutions. Low-resolution projection is then passed sequentially through this stacked pipeline to finally generate the high resolution climate projections.}

\subsubsection{Quantile Mapping Approach (Q-Map)~\cite{cannon2015bias}} Bias Correction Spatial Disaggregation (BCSD) or Quantile mapping approach follows a point based statistical estimation. It is a simple but effective method for statistical downscaling. These models correct systematic biases in distribution of climate variables.


\subsection{Evaluation Metrics}
We leverage root-mean-square error (\textbf{RMSE}), and mean of absolute difference (referred to as \textbf{Bias}) to compare our proposed model with the state-of-the-art frameworks for statistical downscaling of climate variables \textcolor{black}{ResLap,} DeepSD and popularly used Q-Map. 

\subsection{Choice of lag-days}

 While in present studies, we have five temporal inputs with a lag of four, we note that the inclusion of a higher number of temporal variables could result in better performance. However, in this case, our choice of lag is limited by the computing power at our disposal. For this study, we used Nvidia-RTX-2080Ti  with a  batch size of 15. 

\subsection{Comparison}
\label{sec:comparison}
In this work, we compare the performance of the proposed approach with state-of-the-art methods, including \textcolor{black}{ResLap,} DeepSD and Quantile mapping. DeepSD, a convolutional neural network-based super-resolution approach for SD~\cite{deepsd}, recently gained popularity, given its accuracy on a single image super-resolution task. On the other hand, Q-Map~\cite{cannon2011quantile} is a simple yet elegant technique that is widely used for the SD problem.

In our experiments, we apply quantile mapping on daily precipitation to match the probability distribution function of ESM outputs with observations by inverting the cumulative distribution functions. This inversion leads to the removal of systematic bias. Since the spatial resolution of ESM and observations is different, we interpolate ESM to common grid resolution to match the observation grid. We note that while the grid size is reduced, mere interpolation does not account for increased resolution as fine-scale information is missing from the interpolated image. The process of interpolation is followed by determining the scaling factors to map the distribution of interpolated ESM outputs to observations. The projections for the year 2000--2005 are used for comparison to BCSD. 

\textcolor{black}{Secondly,  ResLap is applied to downscale over India using coarse resolutions from ESM and it's corresponding observations. ResLap consists of residual dense blocks(RDB) \cite{rdbnet} embedded into the Laplacian pyramid super-resolution network (LapSRN) \cite{laplacenet} with the addition of Charbonnier function as the loss metric.}

Lastly, we apply DeepSD using elevation and coarse-resolution precipitation obtained from ESM as input channels to learn the non-linear mapping between the ESM projection and precipitation. We use an super-resolution Convolution Neural Network (SRCNN) architecture, as presented by~\cite{dong2014learning} and~\cite{vandal2017deepsd}.

\section{Results and Evaluation}
\label{sec:results}
{ We analyse and compare the performance of our proposed method with the other methods described in Section \ref{sec:comparison}. Key metrics such as Root mean square error (RMSE) and Bias are used to capture the predictive capabilities of these methods. Secondly, We compare the performance of these methods on extreme precipitation over India to analyse the applicability of the methods on extreme events.}
\begin{figure*}[!tbh]
        \centering 
        \includegraphics[width=2\columnwidth]{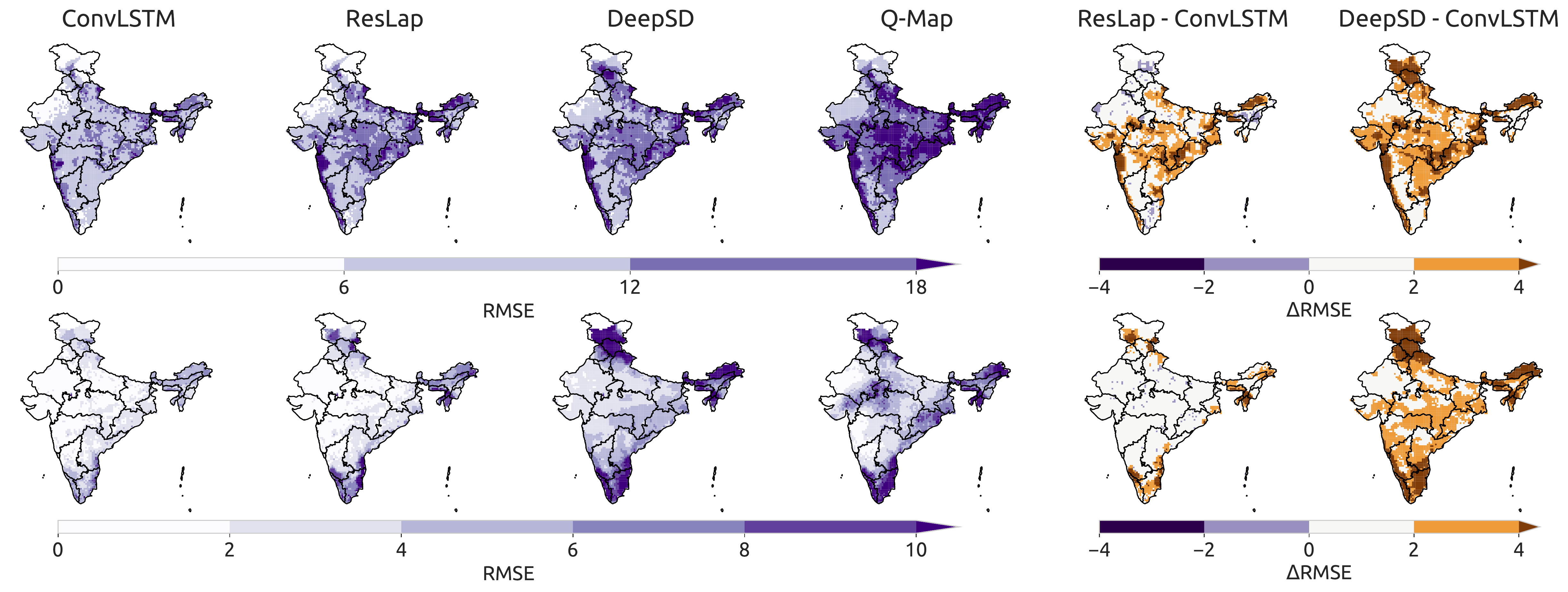} 
        \caption[a floating figure]{\textbf{Daily root-mean-square error (RMSE) computed for (Top) monsoon period and (Bottom) non-monsoon period between years 2000--2005 (test set) for \textcolor{black}{ConvLSTM, ResLap and DeepSD.} The right figure depicts the difference in RMSE values of \textcolor{black}{ConLSTM with ResLap and DeepSD} on the same period. It can be observed that \textcolor{black}{both ResLap and DeepSD}  over predicts on most grid points even during monsoon period.}}
        \label{fig:lowobs} 
        \label{fig:highobs}
\end{figure*}

\begin{figure*}[!tbh]
        \centering 
        \includegraphics[width=2\columnwidth]{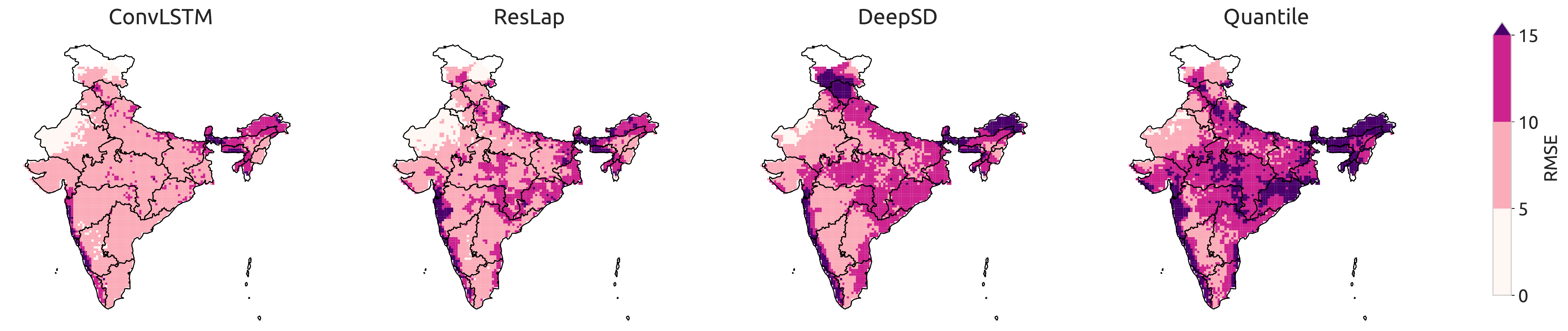} 
        \caption[a floating figure]{\textbf{(Best viewed in color) Daily root-mean-square error (RMSE) computed for years between 2000--2005 (test set) for ConvLSTM, \textcolor{black}{ResLap,} DeepSD and Q-Map.}}
        \label{fig:fullobs} 
\end{figure*}
\begin{figure*}[!tbh]
        \centering 
        \includegraphics[width=2\columnwidth]{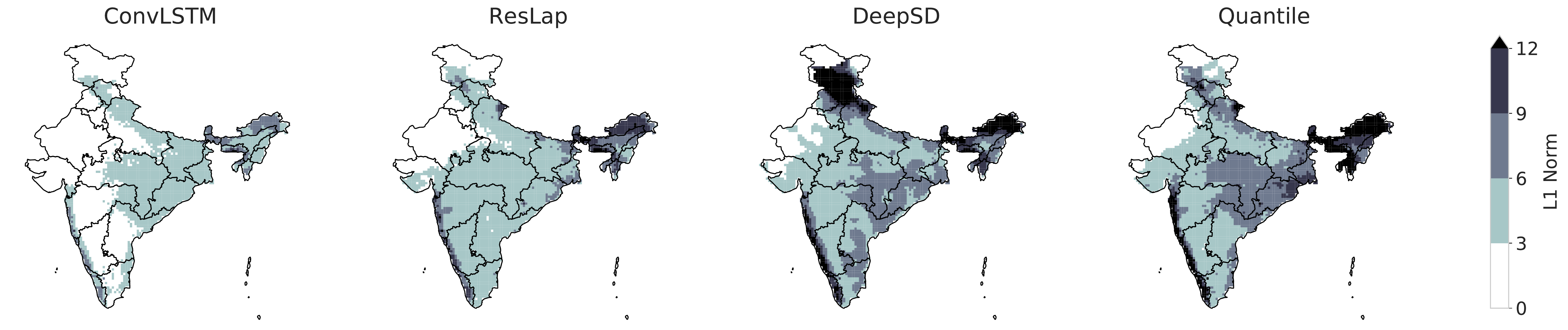} 
        \caption[a floating figure]{\textcolor{black}{\textbf{(Best viewed in color) L1 Norm computed on average projection for years between 2000--2005 (test set) for ConvLSTM, ResLap, DeepSD and Q-Map.}}}
        \label{fig:fullobs} 
\end{figure*}

\subsection{Daily Predictability}
\textcolor{black}{Results of Root mean square error (RMSE) of the predicted projections of precipitation over India during the months of monsoon and non-monsoon is shown in \textbf{Fig~\ref{fig:lowobs}}. The difference in RMSE values of our method with that of ResLap and DeepSD over each location in India is given on the right-side in \textbf{Fig~\ref{fig:lowobs}}. The lower rate of errors of our method on both monsoon and non-monsoon period over India shows the improved performance of ConvLSTM. Similarly, \textbf{Fig~\ref{fig:fullobs}} shows the projection of overall Root mean square error (RMSE) of the methods over the whole test set of precipitation (\textit{2000-2005}).} It can be clearly observed that despite ESM having false projections over East-part of India, the model generates more accurate projections over the southern part of India along with maintaining the extreme values.



The predictive performance of the model in terms of root-mean-square-error (RMSE) and Bias values on monsoon and non-monsoon are shown in \textbf{Table~\ref{tab:eval1}}. The model has performed remarkably well in the non-monsoon periods with a test accuracy of $2.37mm$, and even with the Monsoon period, model has performed well enough with test RMSE of $9.86mm$ despite the period having more extremes over the southern region with a large offset between the ESM and the observed data. \textbf{Table~\ref{tab:eval2}} shows the overall predictive performance of the models on the whole test set (\textit{2000-2005}). \textcolor{black}{\textbf{Table~\ref{tab:season}} shows a season-wise comparison of  RMSE and Bias values. Again, it can be observed clearly that ConvLSTM outperforms the other methods for chosen metrics in all seasons.}
\begin{table}[!tbh]
    \caption{\textbf{Performance of our proposed model against \textcolor{black}{ResLap,} DeepSD and Q-Map on the period of 2000-2005 (test data) for Monsoon and non-monsoon period}}    
    \centering
    \label{TestTable}

            \centering
        \begin{tabular}[t]{lcccccc}
        \toprule
        &&Our&ResLap&DeepSD&Q-Map\\
        [0.5ex]
        \hline
        \noalign{\vskip 0.1cm}
        \multirow{2}{*}{\parbox{1.5cm}{Non Monsoon}}
                                  &RMSE& \textbf{2.37} &4.02& 5.77 &7.04 \\
                                  &Bias& \textbf{0.79} &1.23& 4.18 &1.94\\
        
        \hline
        \noalign{\vskip 0.1cm}
        \multirow{2}{*}{\parbox{1.5cm}{Monsoon}}
                              &RMSE& \textbf{9.86} &12.46& 12.66& 19.48\\
                              &Bias& \textbf{5.18} &6.23& 7.15 & 9.99&\\
        \bottomrule

        \end{tabular}
    \label{tab:eval1}
\end{table}

\begin{table}[!tbh]
    \caption{\textbf{Overall Performance of our proposed model against \textcolor{black}{ResLap,} DeepSD and Q-Map on the period of 2000-2005 (test data) in India.}}
            \centering
        \begin{tabular}{ccccc} 
        \toprule
         & Our & ResLap & DeepSD & Q-Map \\ 
        [0.5ex] 
        \hline 
        \noalign{\vskip 0.1cm}
        RMSE &\textbf{7.94}&9.88&11.42&14.16\\
        Bias &\textbf{3.17}&3.72&5.97& 4.63\\
        \bottomrule
        \end{tabular}
        \label{tab:eval2}
\end{table}%

\begin{table}[!tbh]
    \caption{\textbf{Season-wise Performance of our proposed model against ResLap, DeepSD and Q-Map on the period of 2000-2005 (test data) in India.}}
    \centering 
    \begin{tabular}{cccccc} 
    \toprule
     && Our & ResLap & DeepSD & Q-Map \\ 
    [0.5ex] 
    \hline
    \noalign{\vskip 0.1cm}
    \multirow{4}{*}{\rotatebox[origin=c]{90}{\centering\textbf{RMSE}}}    
    &DJF &\textbf{2.38} & 3.31 & 6.69 & 6.45\\
    &JJA &\textbf{12.78}& 15.62 & 16.74 & 21.68\\
    &MAM &\textbf{4.03} & 5.07 & 7.30 & 8.15\\
    &SON &\textbf{8.32} & 10.39 & 11.92 & 14.36\\
    \hline
    \noalign{\vskip 0.1cm}
    \multirow{4}{*}{\rotatebox[origin=c]{90}{\centering\textbf{Bias}}}    
    &DJF &\textbf{0.64} & 1.15 & 4.07 & 0.98\\
    &JJA &\textbf{6.298}& 7.65 & 8.49 & 10.06\\
    &MAM &\textbf{1.36} & 1.60 & 4.21 & 1.94\\
    &SON &\textbf{3.61} & 4.45 & 6.60 & 5.12\\
    \bottomrule
    \end{tabular}
\label{tab:season}
\end{table}

\subsection{Predicting Extremes}

\begin{figure}[t]
    \centering
    \includegraphics[width=0.4\textwidth]{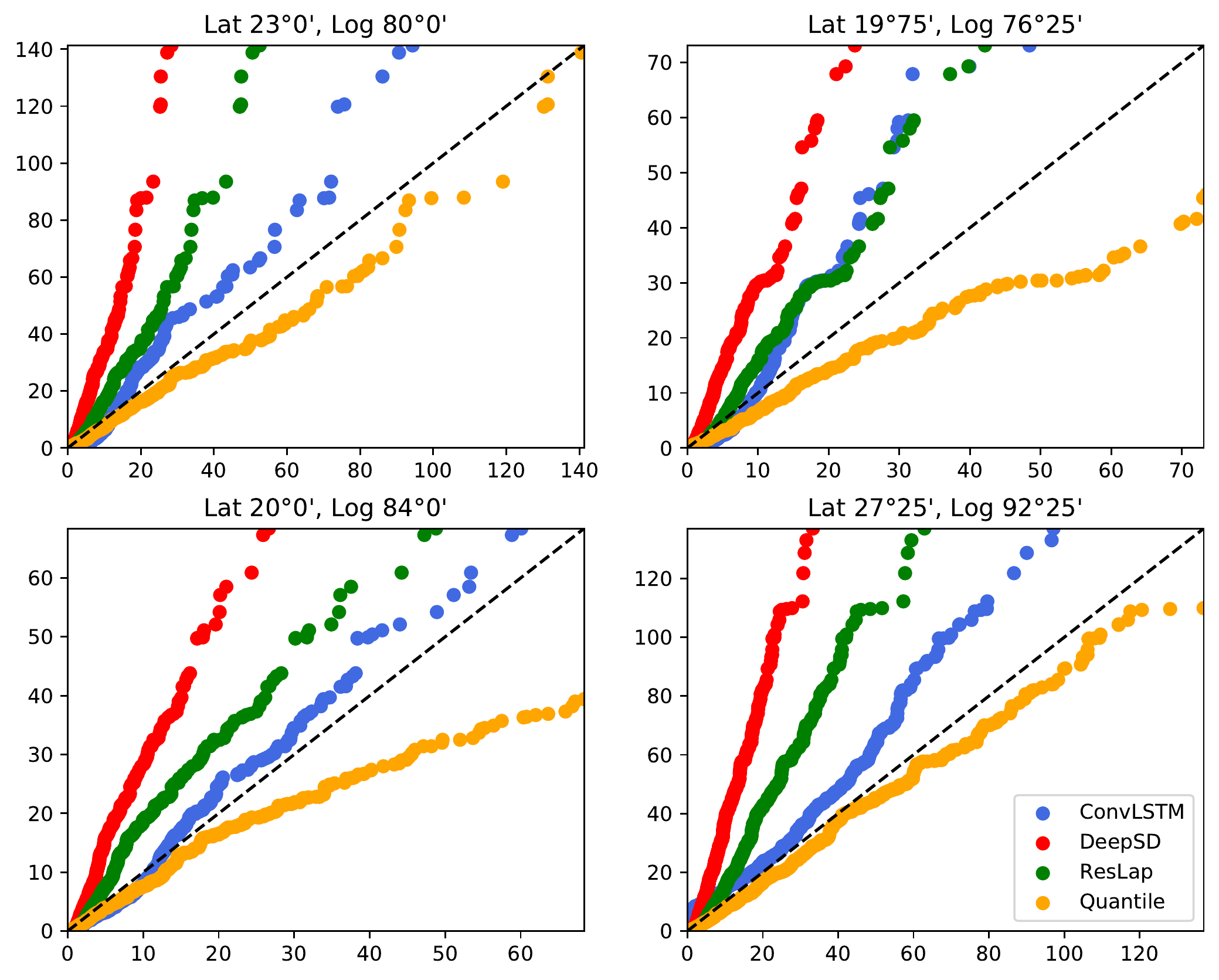}
    \caption{\textcolor{black}{\textbf{Quantile vs Quantile (Q-Q) plots on 4 randomly chosen grid points over India. The x-axis denotes the actual expected quantiles and y-axis denotes the quantiles produced by the methods.}}}
    \label{fig:qq_random}
\end{figure}

\begin{figure}[t]
    \centering
    \includegraphics[width=0.4\textwidth]{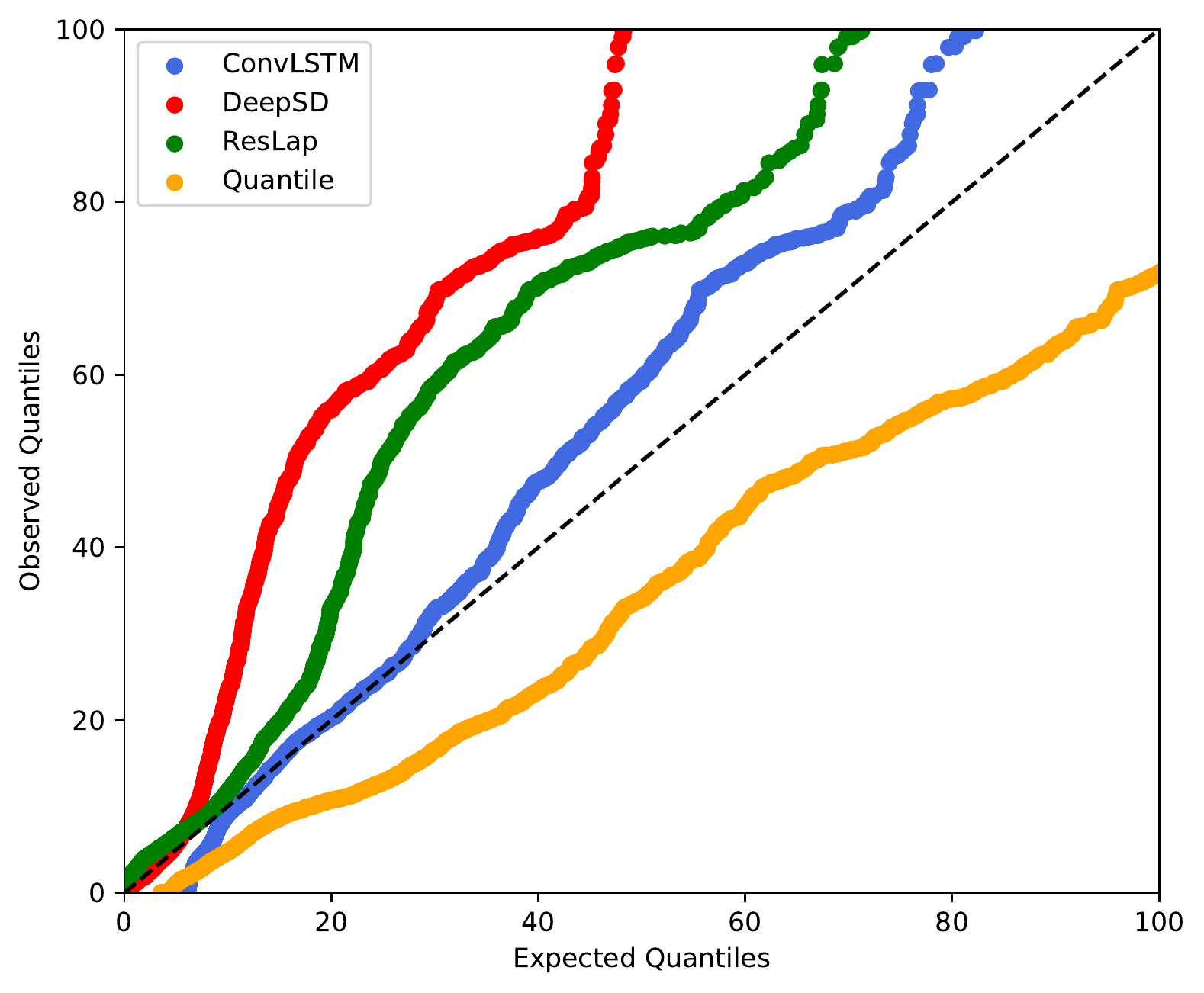}
    \caption{\textcolor{black}{\textbf{Quantile vs Quantile (qq) plot of the projection of grid points averaged over test period on India. The x-axis denotes the actual expected quantiles and y-axis denotes the quantiles produced by the methods.}}}
    \label{fig:qq_avg}
\end{figure}

\begin{figure*}[t]
        \centering 
        \includegraphics[width=2 \columnwidth]{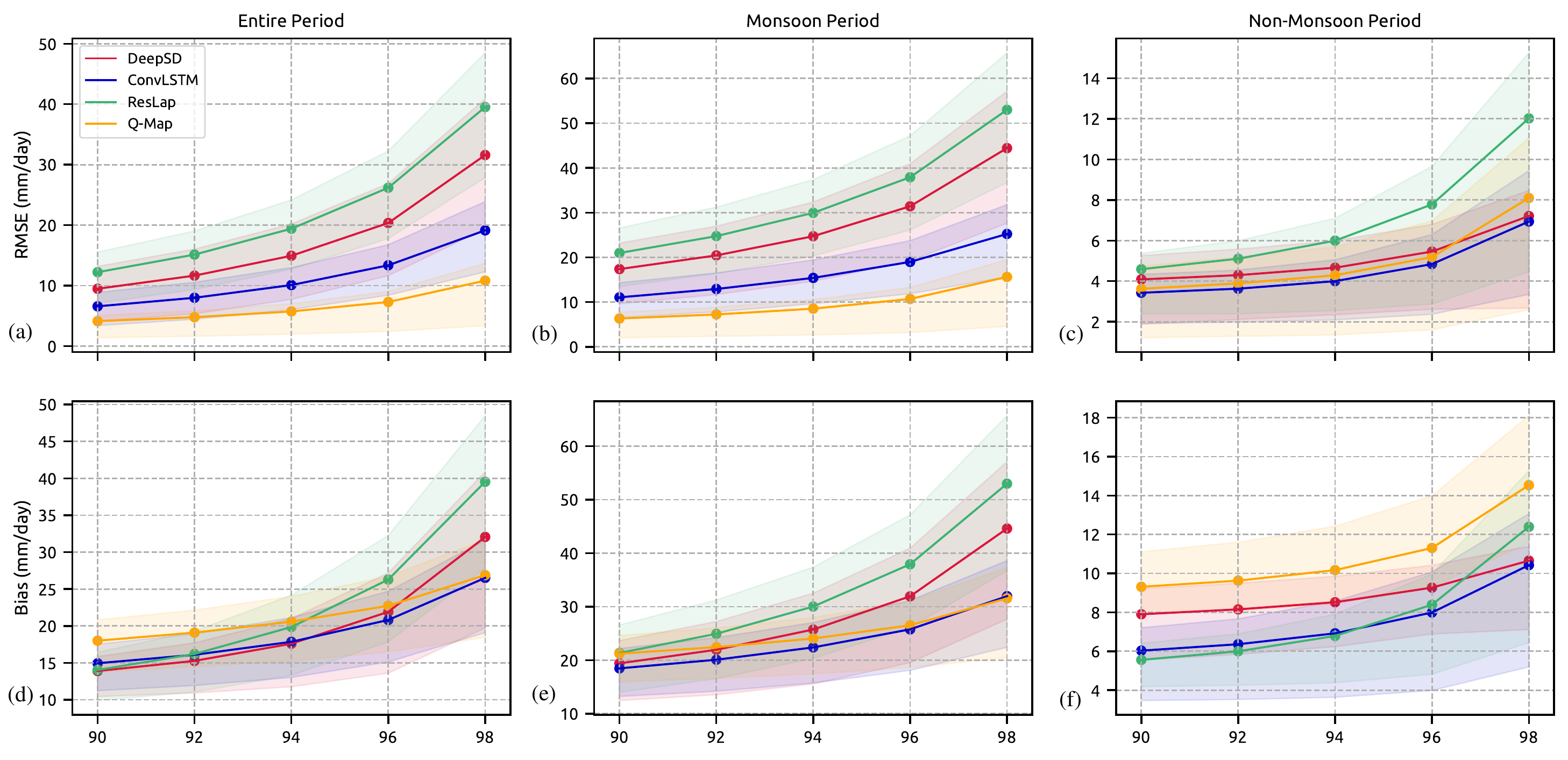}  
        \caption[a floating figure]  {\textbf{(Best viewed in color) 
        \textcolor{black}{Comparison of predictive performance on extreme precipitation of ConvLSTM, ResLap and DeepSD on the entire (a, d), monsoon (b, e) and non-monsoon (c, f) periods betweeen the year 2000-2005. At each location in India, all precipitation events above a percentile threshold (x-axis) are selected. Percentile thresholds between 90 and 99.9 are used. RMSE (a, b, c) and Bias (d, e, f) are computed for these events over India and averaged over the period for years 2000 to 2005 (test data). Confidence intervals show inter quantile range computed over entire India.}}}
        \label{fig:plot2} 
\end{figure*}

\textcolor{black}{We perform various evaluations of the methods on extreme events of precipitation. For this, first we consider the precipitation events that are greater than the $90$th percentile for each location over the grid of India. We select the $25$th and the $75$th quantiles of both RMSE and Bias values for determining the confidence intervals. \textbf{Fig~\ref{fig:plot2}} shows the plot of mean along with the computed confidence bounds of Root mean square error (RMSE) and Bias values on the percentile threshold between $90$ and $98$. It can be seen clearly that both ResLap and DeepSD overpredicts the precipitation values at the extremes and performs the worst. Q-map manages to lower the RMSE value on extremes for the monsoon-period, however, it is observed that it's performance is not stable and produces high biased values on all the other periods. Even though RMSE and Bias values deviates much more on the extremes, however, ConvLSTM is able to maintain stable performance and overall still outperforms the other methods. \textcolor{black}{We also construct quantile-quantile(Q-Q) plots to measure the overall performance of methods on extreme events. We select four random grid-points over India and we compare the quantiles generated by the methods with the quantile of the observed precipitation. \textbf{Fig~\ref{fig:qq_random}} shows the Q-Q plots on the four randomly chosen grid points. We also contruct Q-Q plot of the entire grid of India averaged over the test period. \textbf{Fig~\ref{fig:qq_avg}} shows this average Q-Q plot. It can be observed that the quantiles of both DeepSD~\cite{deepsd} and ResLap~\cite{reslap} deviates from the observation. The quantile produced by Q-map is closer for some grid points but it is has large deviation on other points. Our approach on the other hand, produces stable quantiles and on average, it has the closest quantiles with that of the observed projection.} }


\section{Conclusion}
\label{sec:con}
While state-of-the-Art Super-resolution architectures have shown promise for single image SD, ESMs models have complex spatiotemporal dependencies that need to be accounted for explicitly. Second, Due to a high computational cost associated with ConvLSTMs, the effective sequence length (the number of days to be included as lag variables) that could be used as an input is limited by the computing resources available. Hence, future dependencies need to account for long term spatiotemporal variability in addition to near-term dependencies, which is a typical characteristic of climate systems that shape up the atmospheric processes. Finally, handling non-stationarities (both in space and time) remains an open question in the field of SD. Future work needs to build on design-of-experiments strategies proposed by Salvi et. al ~\cite{salvi2016credibility} to evaluate the performance of SD approaches under non-stationary climate scenarios. We emphasize that while there has been growing interest of the scientific community in the applications of deep learning architectures in generating high-resolution outputs of Earth System Models, translation of these outputs to design relevant intensity, duration and frequency curves require more robust evaluation metrics of extremes which are stakeholder relevant. 

\section*{REPRODUCIBILITY}
We have utilized only free open-source scientific libraries and frameworks for the experiments. The coarse resolution precipitation data from NCAR Community Earth System Model (NCAR CESM1 CAM5) is available on climate model inter-comparison project (CMIP5) archives and auxilliary variables used as part of the study are available from UCAR NCAR repository. All codes used to generate these results are available on Github \footnote{\url{https://github.com/cryptonymous9/Augmented-ConvLSTM}}.

\section*{Acknowledgment}
High-resolution observations over India were distributed by Indian Meteorological Department located at Pune. We thank Professor Auroop Ganguly for valuable comments and suggestions. We thank Dr. Thomas Vandal for sharing codes necessary to reproduce DeepSD results.

\bibliographystyle{ACM-Reference-Format}
\bibliography{sample-base}
\flushend

\end{document}


\title{[Supplementary] Augmented Convolutional LSTMs for Generation of High-Resolution Climate Change Projections}

\author{Nidhin Harilal}
\affiliation{%
  \institution{IIT Gandhinagar}}
\email{nidhin.harilal@iitgn.ac.in}

\author{Udit Bhatia}
\affiliation{
  \institution{IIT Gandhinagar}}
\email{bhatia.u@iitgn.ac.in}

\author{Mayank Singh}
\affiliation{
  \institution{IIT Gandhinagar}}
\email{singh.mayank@iitgn.ac.in}

\renewcommand{\shortauthors}{Harilal, et al.}
\maketitle

\section*{REPRODUCIBILITY}
We have utilized only free open-source scientific libraries and frameworks for the experiments. The coarse resolution precipitation data from NCAR Community Earth System Model (NCAR CESM1 CAM5) is available on climate model inter-comparison project (CMIP5) archives and auxilliary variables used as part of the study are available from UCAR NCAR repository. All codes used to generate these results are available on Github.
Projection of changes in extreme indices of climate variables suchas temperature and precipitation are critical to assess the potentialimpacts of climate change on human-made and natural systems,including critical infrastructures and ecosystems. While impactassessment and adaptation planning rely on high-resolution pro-jections (typically in the order of a few kilometers), state-of-the-artEarth System Models (ESMs) are available at spatial resolutionsof few hundreds of kilometers. Current solutions to obtain high-resolution projections of ESMs include downscaling approachesthat consider the information at a coarse-scale to make predictionsat local scales. Complex and non-linear interdependence amonglocal climate variables (e.g., temperature and precipitation) andlarge-scale predictors (e.g., pressure fields) motivate the use of neu-ral network-based super-resolution architectures. In this work, wepresent auxiliary variables informed spatio-temporal neural archi-tecture for statistical downscaling. The current study performsdaily downscaling of precipitation variable from an ESM output at1.15 degrees ( 115 km) to¼degrees (25 km) over the world’s mostclimatically diversified country, India. We showcase significant im-provement gain against two popular state-of-the-art baselines witha better ability to predict extreme events. To facilitate reproducibleresearch, we make available all the c